%% file: main.tex
\documentclass[runningheads]{llncs}

\usepackage{graphicx}
\usepackage{bbm}
\usepackage{amsmath}
\usepackage{amsfonts}
\usepackage[misc]{ifsym}
\begin{document}

\title{Polyp-artifact relationship analysis using graph inductive learned representations}

\titlerunning{polyp-artifact relationship analysis using graph inductive learning}

\author{Roger D. Soberanis-Mukul\inst{1}\textsuperscript{(\Letter)} \and
Shadi Albarqouni\inst{1,3}\and
Nassir Navab\inst{1,2}}

\authorrunning{R.D. Soberanis-Mukul, S. Albarqouni, N. Navab}

\institute{Computer Aided Medical Procedures, Technische Universit\"{a}t M\"{u}nchen, Germany 
\email{roger.soberanis@tum.de} \and
Computer Aided Medical Procedures, Johns Hopkins University, Baltimore, USA \and
Computer Vision Laboratory, ETH Zurich, Switzerland }

\maketitle 

\begin{abstract}
The diagnosis process of colorectal cancer mainly focuses on the localization and characterization of abnormal growths in the colon tissue known as polyps. Despite recent advances in deep object localization, the localization of polyps remains challenging due to the similarities between tissues, and the high level of artifacts. Recent studies have shown the negative impact of the presence of artifacts in the polyp detection task, and have started to take them into account within the training process. However, the use of prior knowledge related to the spatial interaction of polyps and artifacts has not yet been considered. In this work, we incorporate artifact knowledge in a post-processing step. Our method models this task as an inductive graph representation learning problem, and is composed of training and inference steps.  Detected bounding boxes around polyps and artifacts are considered as nodes connected by a defined criterion. The training step generates a node classifier with ground truth bounding boxes. In inference, we use this classifier to analyze a second graph, generated from artifact and polyp predictions given by region proposal networks. We evaluate how the choices in the connectivity and artifacts affect the performance of our method and show that it has the potential to reduce the false positives in the results of a region proposal network.

\end{abstract}

\input{introduction}
\input{methods}
\input{experiments_results}

\section*{Acknowledgments}
R. D. S. is supported by Consejo Nacional de Ciencia y Tecnolog\'{i}a (CONACYT), Mexico. 
S.A. is supported by the PRIME programme of the German Academic Exchange Service (DAAD) with funds from the German Federal Ministry of Education and Research (BMBF).

\bibliographystyle{splncs04}
\bibliography{bib.bib}

\end{document}

%% file: introduction.tex
\section{Introduction}

\begin{figure}[t]
\begin{center}
\includegraphics[width=0.9\textwidth]{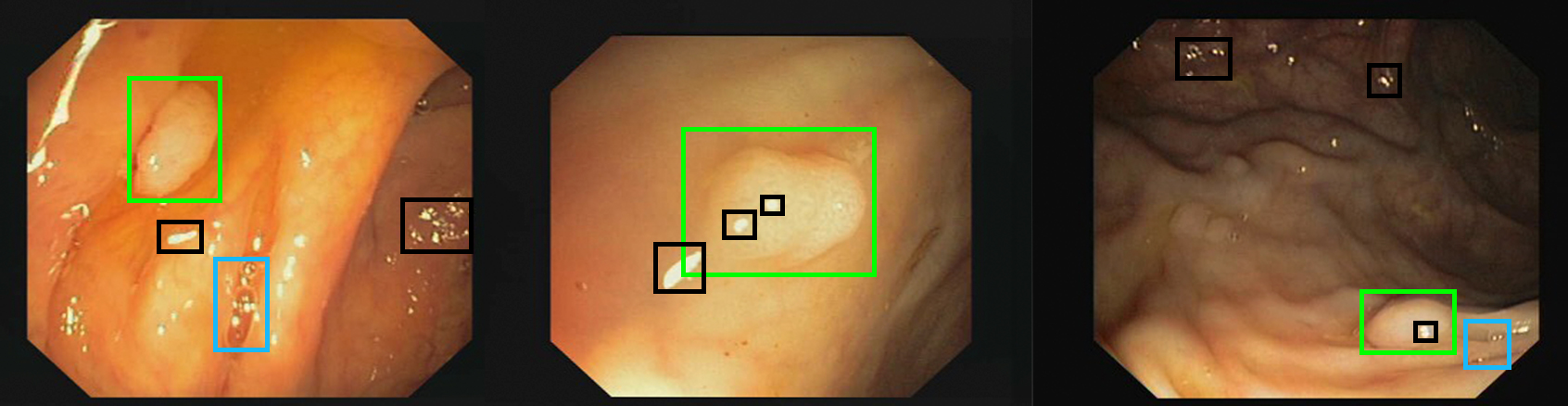}    
\end{center}
\caption{Best viewed in color. Polyps samples (green boxes), bubbles samples (blue), and specularities. Samples from the CVC\_ClinicDB dataset.} \label{fig:samples}
\end{figure}

One of the main steps in the diagnosis of colorectal cancer (CRC) is the detection and characterization of polyps (Fig. \ref{fig:samples}). Colorectal polyps are abnormal growths in the colon tissue, and they can be either benign or develop into CRC. The size and type of the polyps are
key indicators in the diagnosis of this disease. For this, the endoscopic analysis of the colon (colonoscopy) is a common diagnostic procedure \cite{bib:chadebecq2015}. Early detection of polyps contributes to lowering the risk for this disease \cite{bib:sornapudi19}. This has motivated the development of computer-aided methods for polyp detection, that initially relied on handcrafted features \cite{bib:gross2009,bib:ganz2012,bib:Bernal2012}. Also, different polyps datasets have been released. As an example, the polyp localization challenge, proposed during MICCAI 2015 \cite{bib:bernal2017}, provided different sets of still frames (CVC\_ClinicDB dataset \cite{bib:BERNAL20159}, ETIS-Larib dataset \cite{bib:silva2014}, and ASU-Mayo Clinic Colonoscopy Video (c) Database \cite{bib:tajbakhsh2015})  for the tasks of precise polyp localization and polyp detection. 

Recent models \cite{bib:qadir2019,bib:shin2018} are motivated by the advances in deep learning for object localization, where region proposal neural networks (RCNN) like Faster RCNN \cite{bib:ren2015} and RetinaNet \cite{bib:lin2017} are employed. The use of  RCNNs in the medical domain still challenging. In the polyp localization problem, the methods should work on a highly dynamic environment, with similarities between tissues \cite{bib:shin2018}. Also, the presence of different endoscopic artifacts like blur,  bubbles, and specularities (Fig. \ref{fig:samples}) have an impact on the model's performance \cite{bib:Bernal2012,bib:bernal2017,bib:soberanis2020}.  
Artifacts are an important component in endoscopy image analysis. Initial works have proposed methods for their identification and removal \cite{bib:stehle2006,bib:funke2018}. Other authors have evaluated artifact impact on the model's performance \cite{bib:bernal2017,bib:vazquez2017,bib:soberanis2020}, and the effects of incorporate artifact class information in the training of multi-class and multi-task approaches \cite{bib:vazquez2017,bib:soberanis2020}. Recently, the endoscopic artifact detection challenge (EAD), has drawn the attention to the role of artifacts in the analysis of endoscopic sequences \cite{bib:ali20191,bib:ali20192,bib:ali2020}, remarking its importance in the definition of models for endoscopy image analysis. 

Similar to the proposal in \cite{bib:soberanis2019} for graph-based segmentation refinement, we propose a graph post-processing step to analyze the polyp-artifact relationship. For this, we define a graph with polyp and artifact bounding boxes. The bounding boxes are connected considering overlap and class similarity. We aim to reduce the false positives for polyp predictions by analysing their relationship with artifact bounding boxes. For this, we train a node classifier using GraphSAGE \cite{bib:Hamilton2017} on a graph generated with ground truth bounding boxes. At inference time, we use standard RCNN models to predict a set of artifact and polyp bounding boxes in the testing dataset. Then, we define a new unseen graph with these predictions. Finally, we use the node classifier to obtain new class predictions for the bounding boxes in the unseen graph. 

Our \textbf{contributions} are twofold: first, to the best of our knowledge, this is the first application of inductive graph learning in the medical problem of polyp localization; and second, this is one of the first works that consider artifact influence in polyp localization, from the spatial point of view.  

%% file: methods.tex
\section{Method}

\begin{figure}[t]
\begin{center}
\includegraphics[width=\textwidth]{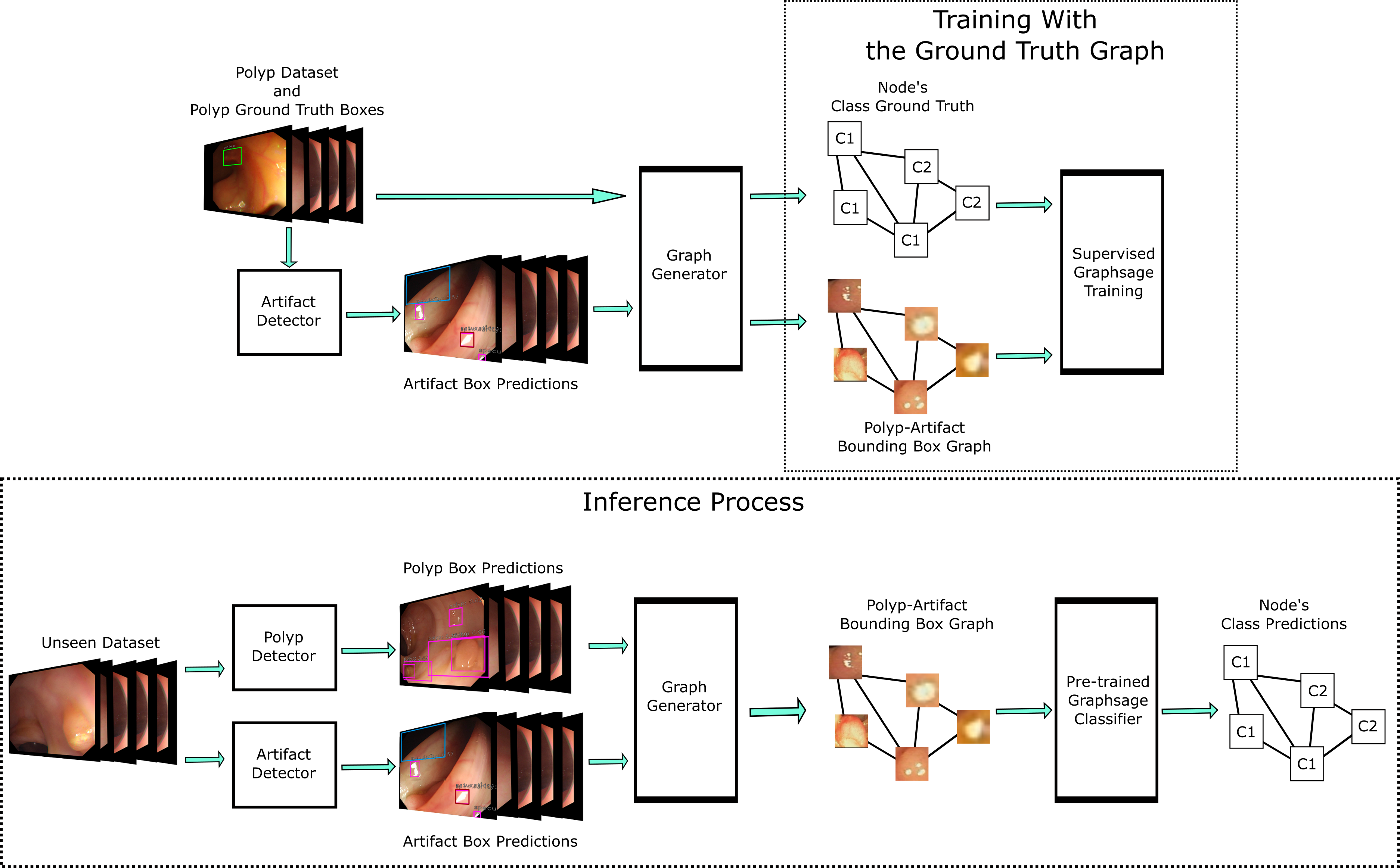}
\end{center}
\caption{Top: Training process. A training dataset with polyps ground truth and artifact predictions are employed to generate a training graph. This is used to train a node classifier using supervised GraphSAGE \cite{bib:Hamilton2017}. Bottom: Inference process. For an unseen dataset, a graph is generated with the RCNN predictions. Then, new class predictions are obtained for the bounding boxes, with the trained node classifier.} \label{fig:framework}
\end{figure}

Our method defines a graph over artifact and polyp bounding boxes. In this section, we describe the graph definition and the training and inference process of the node classifier (Figs. \ref{fig:links} and \ref{fig:framework}).

\subsection{Graph of Bounding Boxes} \label{subsec:graph_bboxes}

\begin{figure}[t]
\begin{center}
\includegraphics[width=0.75\textwidth]{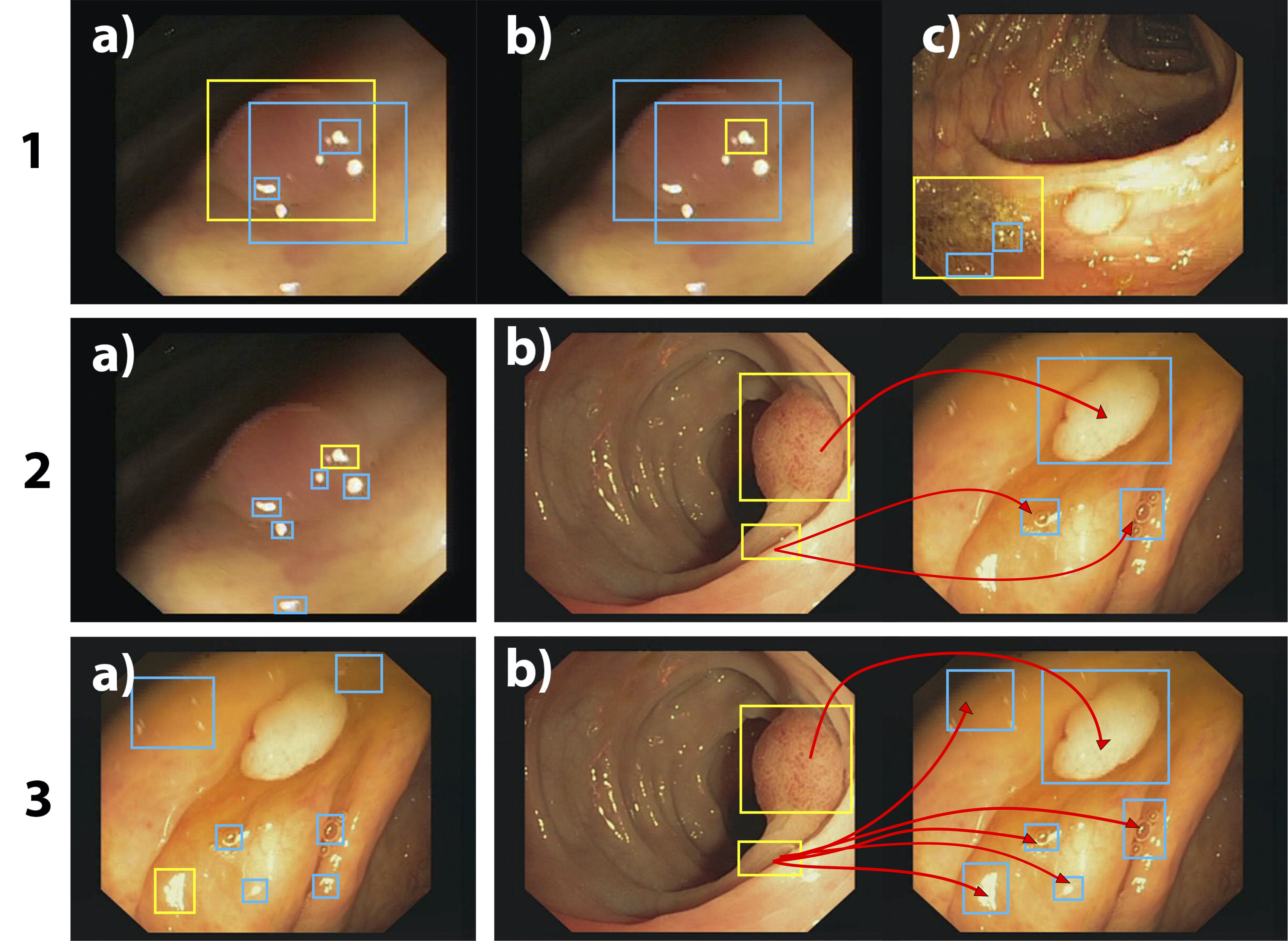}    
\end{center}
\caption{Connectivity criteria (best viewed in color. Yellow boxes indicate a particular (source) bounding box. Examples of its neighbors (destination) bounding boxes are in blue. Red arrows are inter-frame connections. Row 1: in-frame overlap connection, with examples of 1a) polyp to artifacts, 1b) artifact to polyp and, 1c) artifact to artifact connections. Row 2: class similarity  connections 2a) in-frame  and 2b) inter-frame. Row 3: binary connectivity, 3a) in-frame and 3b) inter-frame. 
} \label{fig:links}
\end{figure}

For both, the training and inference process, our method requires the definition of a graph $\mathcal{G}(\mathcal{V}, \mathcal{E})$. Consider a given set of $n$ endoscopic still frames $D = \{I_1, I_2, ..., I_n\}$. For each still frame $I_i$, we have a set of polyp bounding boxes $Y_i = \{y_{1}^i, ..., y_{m_i}^i\}$ and a set of artifact bounding boxes $Z_i = \{z_1^i, ..., z_{k_i}^i\}$ with corresponding types $C_i = \{c_1^i, ..., c_{k_i}^i\}$, with each $y$ and $z$ a parametrization of the bounding boxes, $m_i$ and $k_i$ are the number of polyp and artifacts, respectively, in frame $I_i$, and $c_j\in$ class(artifact) indicates the class of the artifact contained in the bounding box $z_j$. 

The nodes of $\mathcal{G}$ are defined by the union of all the bounding boxes in set D, this is:
\begin{equation}
\mathcal{V} = \bigcup\limits_{i=1}^{n} Y_i \cup Z_i
\end{equation}
Each node $x_i \in \mathcal{V}$ will have a unique class $c'_i \in$ class(artifact) $\cup$ \{`polyp'\}. The set of links  $(x_i, x_j) \in \mathcal{E}$, $x_i, x_j \in \mathcal{V}$ is defined based on a set of conditions.
We propose different combinations of four connectivity criteria. These criteria are motivated by the ideas that boxes with the same or similar classes should be connected, and artifacts overlapping with polyps can influence the detection results.
\begin{enumerate}
    \item We connect two bounding boxes with a probability $p=0.5$. 
    \item Two boxes are connected independently of their class if they overlap with an $IoU > 0.5$ or if one bounding box is contained inside the other (Fig. \ref{fig:links}, row 1)).
    \item Two bounding boxes are connected if they have the same class (class similarity, see Fig. \ref{fig:links}, row 2).
    \item Two bounding boxes are connected if they are both polyps or if they are both artifacts, independently of the artifact class (binary, see Fig. \ref{fig:links}, row 3). 
\end{enumerate}

All criteria are symmetric and self-connections are not allowed. We can differentiate between connections in the same frame (in-frame) and between different frames (inter-frame), see Fig. \ref{fig:links}. Criterion 1 can be applied for both connections. Criterion 2 applies only for bounding boxes in the same frame (in-frame). Criteria 3 and 4 are in- and inter-frame, and only one of them can be used at a time to generate the graph.   

\subsection{Graph Training and Inference}

We use GraphSAGE \cite{bib:Hamilton2017} to learn a multi-class node classifier. In this step, the nodes and edges are computed from ground truth bounding boxes of polyps and artifacts. To train, we require both polyps and artifacts to be in the same dataset. Since most datasets only contain polyps, we use an artifact detector to extend a polyp-only dataset. The training process is presented in Fig. \ref{fig:framework}. Note that this process is unrelated to the training of the RCNN models. 

In inference time, we use the node classifier as a post-processing step. Given an artifact RCNN detector $f_a(I)$, and a polyp RCNN detector $f_p(I)$, we predict a set of polyp bounding boxes $Y'_i = f_p(I_i)$ and artifact bounding boxes with classes $Z'_i, C'_i = f_a(I_i)$ for each image $I_i$ in an unseen set $D*$. Then, we use these predictions to construct a new graph, as defined in section \ref{subsec:graph_bboxes}. For connectivity criteria that require to know the bounding box classes, we consider the classes predicted by the RCNNs. Finally, we use the node classifier to generate new bounding box class predictions from a graph-based perspective (Fig. \ref{fig:framework} bottom). 

%% file: experiments_results.tex
\section{Experiments and Results}

\subsection{Datasets}
 
 The CVC\_ClinicDB dataset \cite{bib:BERNAL20159} was used to train both, the RCNN and the GraphSAGE classifier. This dataset contains 612 still frames taken from 29 video sequences. The CVC\_ColonDB dataset \cite{bib:Bernal2012} is reserved for testing. It contains 380 still frames from 15 different studies. Ground truth bounding boxes are extracted from the dataset's segmentation masks. The EAD Challenge's\footnote[1]{https://ead2019.grand-challenge.org/EAD2019/} train dataset is used only for training our artifact detector. It contains annotations for seven different artifacts. 

\subsection{Implementation Details}\label{subsec:imp_details}

The polyp and artifact detectors are both RetinaNet \cite{bib:lin2017} models. For our post-processing method, all the bounding boxes were resized to $64 \times 64$. This allows extracting fix-sized features for all the bounding boxes. The GraphSAGE framework requires feature representation for the nodes. We use histogram of oriented gradients (HOG) as features for the bounding boxes. We use the Pytorch  implementation of GraphSAGE\footnote[2]{https://github.com/williamleif/graphsage-simple/} with the mean aggregator.

For the RCNN models, in all our experiments, we take artifacts with a prediction score $\geq 0.5$. For polyp predictions, we use $0.25$ in order to reduce false negatives. In general, to consider a polyp as detected, we use a condition similar to the MICCAI 2015 polyp detection challenge\footnote[3]{http://endovis.grand-challenge.org}. A polyp is detected if the center of the predicted bounding box is inside a ground truth bounding box. Criteria for false positives (FP), false negatives (FN), and true positives (TP) are taken from the challenge rules. The metrics obtained are precision, recall, F1, and F2 score. We use the F1 score for reference, since it combines both precision and recall. We compare the results against the RCNN output (baseline) before the node-classifier post-processing. We evaluate the effect of different artifact classes, connectivity criteria (cc) and video-level (vl) vs dataset-level (dl) inter-frame connections (ifc). 

\subsection{Artifact Classes}
We evaluate the performance of the node classifier considering different artifact classes. We try with three sets selected according to the analysis presented in \cite{bib:soberanis2020}. To keep the number of samples balanced across classes, we did not include specularities. The first set uses all the EAD artifact's classes available: art1 = \{`polyp', `saturation', `misc.', `blur', `contrast', `bubbles', `instrument'\}. The set art2 = \{`polyp', `misc.', `blur', `bubbles'\}  considers artifacts that are relevant for polyp detection \cite{bib:soberanis2020}. Finally, art3 = art1 - art2 + \{`blur'\}, removes all artifacts in art2, except for blur, since it has show benefits when it is included in the training process of models \cite{bib:soberanis2020}. We use a combination of criteria 2 (overlap) and 3 (same class) for connectivity (see section \ref{subsec:graph_bboxes}). Connection are between bounding boxes in the entire dataset (dataset-level connections).

\begin{table}[t]
			\caption{Performance with different sets of artifact classes. The artifact classes (art. class), connectivity criteria (cc), and inter-frame connections (ifc) are indicated} 
			\label{tab:art_class}
			\centering
			\begin{tabular}{l | c | c | c | c | c | c | c | c | c | c }
			\hline\noalign{\smallskip}
			model & art. class & cc & ifc & TP & FP & FN & Precision & Recall & F1 & F2   \\ 
			\noalign{\smallskip}
			\hline
			\hline
			\noalign{\smallskip}
			baseline & n/a. & n/a. & n/a. & 338 & 230 & 42 & 0.595 & 0.889 & {\bf 0.713} & 0.809\\ 
			\noalign{\smallskip} 
			\hline
			\hline
			\noalign{\smallskip} 
			graph 1.1 (ours) & art1 & 2 and 3 & dl & 344 & 370 & 36 & 0.482 & 0.905 & 0.629 & 0.770\\
			\noalign{\smallskip} 
			\hline
			\noalign{\smallskip} 
			graph 1.2 (ours) & art2 & 2 and 3 & dl & 349 & 448 & 31 & 0.438 & 0.918 & 0.593 & 0.753\\
			\noalign{\smallskip} 
			\hline
			\noalign{\smallskip} 
			graph 1.3 (ours) & art3 & 2 and 3 & dl & 340 & 277 & 40 & 0.551 & 0.895 & 0.682 & 0.796\\
			\noalign{\smallskip} 
			\hline
			\end{tabular}
\end{table}
Results are presented in Table \ref{tab:art_class}. The best performance occurs when we do not include the representative artifacts for polyp localization (art3). An explanation can be that those artifacts are easily confused as polyps.  This can also explain the increment in the FP for art2, which only includes these representative artifacts. Art1, which includes both sets, has an intermediate performance.  In any case, the graph was not able to reach the baseline's performance. Our next experiments only consider art1 and art3.

\subsection{Binary Connectivity}
In our next experiment, we change the connectivity criterion 3 (same class) by criterion cc = 4 (binary polyp/artifact), keeping the overlap connectivity (cc = 2). Results are presented in the Table
\ref{tab:connectivity}. Considering all artifacts into the same class shows a benefit in the performance of the graph, and perform slightly better than the baseline. Considering all artifacts into the same class shows a benefit in the performance of the graph.  This can also reduce the impact of errors in the artifact RCNN. The RCNN could predict the wrong individual class for an artifact bounding box, but it still is part of the general `artifact' class.  To verify that the our connectivity gives us significant information, we try our framework with a graph generated with random connections (connectivity criterion 1 in our list). Here, two different bounding boxes will be connected with a probability of $p=0.5$. Doing this, the performance drops to $0.49$. This shows that our connectivity strategies give useful information to the graph, compared with random connections. 

\begin{table}[t]
			\caption{Performance with connectivity criteria 2 and 4.The artifact classes (art. class), connectivity criteria (cc), and inter-frame connections (ifc) are indicated} 
			\label{tab:connectivity}
			\centering
			\begin{tabular}{l | c | c | c | c | c | c | c | c | c | c }
			\hline\noalign{\smallskip}
			model & art. class & cc & ifc  & TP & FP & FN & Precision & Recall & F1 & F2   \\  
			\noalign{\smallskip}
			\hline
			\hline
			\noalign{\smallskip}
			baseline & n/a. & n/a. & n/a. & 338 & 230 & 42 & 0.595 & 0.889 & 0.713 & 0.809\\ 
			\noalign{\smallskip} 
			\hline
			\hline
			\noalign{\smallskip} 
			graph 2.1 (ours) & art1 & 2 and 4 & dl & 318 & 167 & 62 & 0.656 & 0.837 & {\bf 0.735} & 0.793\\
			\noalign{\smallskip} 
			\hline
			\noalign{\smallskip} 
			graph 2.2 (ours) & art3 & 2 and 4 & dl & 330 & 210 & 50 & 0.611 & 0.868 & 0.717 & 0.801\\
			\noalign{\smallskip} 
			\hline
			\noalign{\smallskip} 
			graph 2.3 (ours) & art1 & 1 & dl & 350 & 696 & 30 & 0.335 & 0.921 & 0.491 & 0.682\\
			\noalign{\smallskip} 
			\hline
			\end{tabular}
\end{table}

\subsection{Video-level Connections}

Now, we kept the configurations of `graph 2.1' and `graph 2.2' in Table \ref{tab:connectivity}, but this time, we generate connections only between frames of the same video (vl). The results are presented in Table \ref{tab:inter_frame}. This has the effect of increase the number of false positives. A reason for this can be that limiting the neighbors to a local neighborhood does not provide enough representative samples to learn the characteristics of polyps and artifacts. 

\begin{table}
			\caption{Performance with video-level connections. The artifact classes (art. class), connectivity criteria (cc), and inter-frame connections (ifc) are indicated} 
			\label{tab:inter_frame}
			\centering
			\begin{tabular}{l | c | c | c | c | c | c | c | c | c | c }
			\hline\noalign{\smallskip}
			model & art. class & cc & ifc  & TP & FP & FN & Precision & Recall & F1 & F2   \\ 
			\noalign{\smallskip}
			\hline
			\hline
			\noalign{\smallskip}
			baseline & n/a. & n/a. & n/a. & 338 & 230 & 42 & 0.595 & 0.889 & {\bf 0.713} & 0.809\\ 
			\noalign{\smallskip} 
			\hline
			\hline
			\noalign{\smallskip} 
			graph 3.1 (ours) & art1 & 2 and 4 & vl & 340 & 481 & 40 & 0.414 & 0.895 & 0.566 & 0.726\\
			\noalign{\smallskip} 
			\hline
			\noalign{\smallskip} 
			graph 3.2 (ours) & art3 & 2 and 4 & vl & 339 & 317 & 41 & 0.517 & 0.892 & 0.654 & 0.779\\
			\noalign{\smallskip} 
			\hline
			\end{tabular}
\end{table}

\subsection{Binary Node Classification}
So far, we have been working with multi-class node classification. For his last experiments, we train the node classification model for binary polyp-artifact classification.  The settings for the graph definitions are the same as models `graph 2.1' and `graph 2.2' in Table \ref{tab:connectivity}. Results in Table \ref{tab:binary} does not show significative differences from Table \ref{tab:connectivity}, except in the number of false negatives for the configuration art1/2,4/y (model `ours 1' in both Tables). The set art1 includes almost all artifact classes. This can cause a class imbalance problem in the binary classification settings, making some polyps being classified as artifacts. 

\begin{table}[t]
			\caption{Node binary classification performance. The artifact classes (art. class), connectivity criteria (cc), and inter-frame connections (ifc) are indicated} 
			\label{tab:binary}
			\centering
			\begin{tabular}{l | c | c | c | c | c | c | c | c | c | c }
			\hline\noalign{\smallskip}
			model & art. class & cc & ifc  & TP & FP & FN & Precision & Recall & F1 & F2   \\ 
			\noalign{\smallskip}
			\hline
			\hline
			\noalign{\smallskip}
			baseline & n/a. & n/a. & n/a. & $338$ & $230$ & $42$ & $0.595$ & $0.889$ & $0.713$ & $0.809$\\ 
			\noalign{\smallskip} 
			\hline
			\hline
			\noalign{\smallskip} 
			graph 4.1 (ours) & art1 & 2 and 4 & dl & 267 & 100 & 113 & 0.728 & 0.703 & 0.715 & 0.707\\
			\noalign{\smallskip} 
			\hline
			\noalign{\smallskip} 
			graph 4.2 (ours) & art3 & 2 and 4 & dl & 326 & 193 & 54 & 0.628 & 0.858 & {\bf 0.725} & 0.799\\
			\noalign{\smallskip} 
			\hline
			\end{tabular}
\end{table}

\section{Conclusion}

Our work presents a novel graph-based representation of polyps and artifact bounding boxes as a potential post-processing step in polyp localization. We have shown that such approaches have the potential to decrease the number false positives of the RCNN prediction, however, with a slightly increment of the false negatives. One limitation of our approach is that it can not recover polyps that were not localized by the RCNN detector. Similarly, the current configuration can only lead to slight improvements over the RCNN. However, we believe that our proposal can lead to future research directions in the artifact-aware polyp localization problem.

%% file: main.bbl
\begin{thebibliography}{10}
\providecommand{\url}[1]{\texttt{#1}}
\providecommand{\urlprefix}{URL }
\providecommand{\doi}[1]{https://doi.org/#1}

\bibitem{bib:ali20191}
Ali, S., Zhou, F., Bailey, A., Braden, B., East, J., Lu, X., Rittscher, J.: A
  deep learning framework for quality assessment and restoration in video
  endoscopy. arXiv preprint arXiv:1904.07073  (2019)

\bibitem{bib:ali2020}
Ali, S., Zhou, F., Braden, B., Bailey, A., Yang, S., Cheng, G., Zhang, P., Li,
  X., Kayser, M., Soberanis-Mukul, R.D., Albarqouni, S., Wang, X., Wang, C.,
  Watanabe, S., Oksuz, I., Ning, Q., Yang, S., Khan, M.A., Gao, X., Realdon,
  S., Loshchenov, M., Schnabel, J., East, J., Wagnieres, G., Loschenov, V.,
  Grisan, E., Daul, C., Blondel, W., Rittscher, J.: An objective comparison of
  detection and segmentation algorithms for artefacts in clinical endoscopy.
  Scientific Reports  (2020)

\bibitem{bib:ali20192}
Ali, S., Zhou, F., Daul, C., Braden, B., Bailey, A., Realdon, S., East, J.,
  Wagni{\`{e}}res, G., Loschenov, V., Grisan, E., Blondel, W., Rittscher, J.:
  Endoscopy artifact detection {(EAD} 2019) challenge dataset. CoRR
  \textbf{abs/1905.03209} (2019), \url{http://arxiv.org/abs/1905.03209}

\bibitem{bib:qadir2019}
Ali~Qadir, H., Shin, Y., Solhusvik, J., Bergsland, J., Aabakken, L.,
  Balasingham, I.: Polyp detection and segmentation using mask r-cnn: Does a
  deeper feature extractor cnn always perform better? In: 3th International
  Symposium on Medical Information and Communication Technology (ISMICT) (2019)

\bibitem{bib:BERNAL20159}
Bernal, J., S\'{a}nchez, F.J., Fern\'{a}ndez-Esparrach, G., Gil, D.,
  Rodr\'{i}guez, C., {n}o, F.V.: Wm-dova maps for accurate polyp highlighting
  in colonoscopy: Validation vs. saliency maps from physicians. Computerized
  Medical Imaging and Graphics  \textbf{43},  99 -- 111 (2015).
  \doi{https://doi.org/10.1016/j.compmedimag.2015.02.007},
  \url{http://www.sciencedirect.com/science/article/pii/S0895611115000567}

\bibitem{bib:Bernal2012}
Bernal, J., S\'{a}nchez, J., {n}o, F.V.: Towards automatic polyp detection with
  a polyp appearance model. In: Pattern Recognition (2012)

\bibitem{bib:bernal2017}
Bernal, J., Tajkbaksh, N., S\'{a}nchez, F.J., Matuszewski, B., Chen, H., Yu,
  L., Angermann, Q., Romain, O., Bj\"{o}rn, Balasingham, I., et~al.:
  Comparative validation of polyp detection methods in video colonoscopy:
  results from the miccai 2015 endoscopic vision challenge. IEEE transactions
  on medical imaging  (2017)

\bibitem{bib:chadebecq2015}
Chadebecq, F., Tilmant, C., Bartoli, A.: How big is this neoplasia? live
  colonoscopic size measurement using the infocus-breakpoint. Medical Image
  Analysis  (2015)

\bibitem{bib:funke2018}
Funke, I., Bodenstedt, S., Riediger, C., Weitz, J., Speidel, S.: Generative
  adversarial networks for specular highlight removal in endoscopic images. In:
  Medical Imaging 2018: Image-Guided Procedures, Robotic Interventions, and
  Modeling. vol. 10576, p. 1057604. International Society for Optics and
  Photonics (2018)

\bibitem{bib:ganz2012}
Ganz, M., Yang, X., Slabaugh, G.: Automatic segmentation of polyps in
  colonoscopic narrow-band imaging data. IEEE Transactions on Biomedical
  Engineering  \textbf{59}(8),  2144--2151 (2012)

\bibitem{bib:gross2009}
Gross, S., Stehle, T., Behrens, A., Auer, R., Aach, T., Winograd, R.,
  Trautwein, C., Tischendorf, J.: A comparison of blood vessel features and
  local binary patterns for colorectal polyp classification. In: Medical
  Imaging 2009: Computer-Aided Diagnosis. vol.~7260, p. 72602Q. International
  Society for Optics and Photonics (2009)

\bibitem{bib:Hamilton2017}
Hamilton, W.L., Ying, R., Leskovec, J.: Inductive representation learning on
  large graphs. In: 31st Conference on Neural Information Processing Systems
  (NIPS 2017) (2017)

\bibitem{bib:soberanis2020}
Kayser, M., Soberanis-Mukul, R.D., Zvereva, A.M., Klare, P., Navab, N.,
  Albarqouni, S.: Understanding the effects of artifacts on automated polyp
  detection and incorporating that knowledge via learning without forgetting
  (2020), arXiv preprint arXiv:2002.02883v3

\bibitem{bib:lin2017}
Lin, T.Y., Goyal, P., Girshick, R., He, K., Doll{\'a}r, P.: Focal loss for
  dense object detection. Proceedings of the IEEE international conference on
  computer vision  (2017)

\bibitem{bib:ren2015}
Ren, S., He, K., Girshick, R., Sun, J.: Faster r-cnn: Towards real-time object
  detection with region proposal networks. In: Advances in neural information
  processing systems (2015)

\bibitem{bib:shin2018}
Shin, Y., Qadir, H.A., Aabakken, L., Bergsland, J., Balasingham, I.: Automatic
  colon polyp detection using region based deep cnn and post learning
  approaches. IEEE Access  (2018)

\bibitem{bib:silva2014}
Silva, J., Histace, A., Romain, O., Dray, X., Granado, B.: Toward embedded
  detection of polyps in wce images for early diagnosis of colorectal cancer.
  International Journal of Computer Assisted Radiology and Surgery
  \textbf{9}(2),  283--293 (2014)

\bibitem{bib:soberanis2019}
Soberanis-Mukul, R.D., Navab, N., Albarqouni, S.: Uncertainty-based graph
  convolutional networks for organ segmentation refinement. In: MIDL (2020)

\bibitem{bib:sornapudi19}
Sornapudi, S., Meng, F., Yi, S.: Region-based automated localization of
  colonoscopy and wireless capsule endoscopy polyps. Appl. Sci.  (2019)

\bibitem{bib:stehle2006}
Stehle, T.: Removal of specular reflections in endoscopic images. Acta
  Polytechnica  \textbf{46}(4) (2006)

\bibitem{bib:tajbakhsh2015}
Tajbakhsh, N., Gurudu, S.R., Liang, J.: Automated polyp detection in
  colonoscopy videos using shape and context information. IEEE transactions on
  medical imaging  \textbf{35}(2),  630--644 (2015)

\bibitem{bib:vazquez2017}
V\'{a}zquez, D., Bernal, J., S\'{a}nchez, F.J., Fern\'{a}ndez-Esparrach, G.,
  L\'{o}pez, A.M., Romero, A., Drozdzal, M., Courville, A.: A benchmark for
  endoluminal scene segmentation of colonoscopy images. Journal of healthcare
  engineering  \textbf{2017} (2017)

\end{thebibliography}
